\documentclass[10pt,journal,compsoc]{IEEEtran}
\ifCLASSOPTIONcompsoc
  \usepackage[nocompress]{cite}
\else
  \usepackage{cite}
\fi
\ifCLASSINFOpdf
   \usepackage[pdftex]{graphicx}
\else
\fi

\usepackage{amsmath}
\usepackage{amsfonts}
\usepackage{algorithm}
\usepackage{algorithmic}
\usepackage{tikz}
\newcommand{\blueline}{
  \raisebox{2pt}{
    \begin{tikzpicture}
      \draw[-,blue,solid,line width = 0.9pt](0,0) -- (5mm,0);
    \end{tikzpicture}
  }
}
\newcommand{\orangeline}{
  \raisebox{2pt}{
    \begin{tikzpicture}
      \draw[-,orange,solid,line width = 0.9pt](0,0) -- (5mm,0);
    \end{tikzpicture}
  }
}

\usepackage{booktabs}

\usepackage{url}

\usepackage{xcolor}

\begin{document}
%
\title{Neural Granger Causality}
%
%
%
%

\urlstyle{tt}

\author{Alex~Tank*,
        Ian~Covert*,
        Nick~Foti,
        Ali Shojaie,
        Emily B. Fox
\IEEEcompsocitemizethanks{
\IEEEcompsocthanksitem * Denotes equal contribution.
\IEEEcompsocthanksitem Alex Tank was with the Department
of Statistics, University of Washington, Seattle, WA, 98103. 
E-mail: alextank@uw.edu
\IEEEcompsocthanksitem Ian Covert, Nicholas Foti, and Emily Fox were with the Department
of Computer Science, University of Washington, Seattle, WA, 98103.
\IEEEcompsocthanksitem Ali Shojaie was with the Department of Biostatistics, University of Washington, Seattle, WA, 98103}
}

\IEEEtitleabstractindextext{%
\begin{abstract}
While most classical approaches to Granger causality detection assume linear dynamics, many interactions in real-world applications, like neuroscience and genomics, are inherently nonlinear. In these cases, using linear models may lead to inconsistent
estimation of Granger causal interactions. We propose a class of nonlinear methods by applying structured multilayer perceptrons (MLPs)
or recurrent neural networks (RNNs) combined with sparsity-inducing penalties on the weights. By encouraging specific sets of weights to
be zero---in particular, through the use of convex group-lasso penalties---we can extract the Granger causal structure. To further contrast
with traditional approaches, our framework naturally enables us to efficiently capture long-range dependencies between series either via
our RNNs or through an automatic lag selection in the MLP. We show that our neural Granger causality methods outperform
state-of-the-art nonlinear Granger causality methods on the DREAM3 challenge data. This data consists of nonlinear gene expression
and regulation time courses with only a limited number of time points. The successes we show in this challenging dataset provide a
powerful example of how deep learning can be useful in cases that go beyond prediction on large datasets. We likewise illustrate our
methods in detecting nonlinear interactions in a human motion capture dataset. 
\end{abstract}

}

\maketitle

\IEEEdisplaynontitleabstractindextext

%
\IEEEpeerreviewmaketitle

\IEEEraisesectionheading{\section{Introduction}\label{sec:introduction}}
In many scientific applications of multivariate time series, it is important to go beyond prediction and forecasting and instead interpret the structure within time series. Typically, this structure provides information about the contemporaneous and lagged relationships within and between individual series and how these series interact. For example, in neuroscience it is important to determine how brain activation spreads through brain regions \cite{sporns2010networks}, \cite{vicente2011transfer}, \cite{stokes2017study}, \cite{sheikhattar2018extracting}; in finance it is important to determine groups of stocks with low covariance to design low risk portfolios \cite{sharpe1998investments}; and, in biology, it is of great interest to infer gene regulatory networks from time series of gene expression levels \cite{fujita2010granger}, \cite{lozano2009groupedtemporal}. However, for a given statistical model or methodology, there is often a tradeoff between the interpretability of these structural relationships and expressivity of the model dynamics.

Among the many choices for understanding relationships between series, Granger causality \cite{granger1969investigating}, \cite{lutkepohl2005new} is a commonly used framework for time series structure discovery that quantifies the extent to which the past of one time series aids in predicting the future evolution of another time series. When an entire system of time series is studied, networks of Granger causal interactions may be uncovered \cite{basu2015network}. This is in contrast to other types of structure discovery, like coherence \cite{sameshima2016methods} or lagged correlation \cite{sameshima2016methods}, which analyze strictly bivariate covariance relationships. That is, Granger causality metrics depend on the activity of the entire system of time series under study, making them more appropriate for understanding high-dimensional complex data streams. Methodology for estimating Granger causality may be separated into two classes, model-based and model-free.

Most classical model-based methods assume linear time series dynamics and use the popular vector autoregressive (VAR) model \cite{lozano2009groupedtemporal}, \cite{lutkepohl2005new}. In this case, the time lags of a series have a linear effect on the future of each other series, and the magnitude of the linear coefficients quantifies the Granger causal effect. Sparsity-inducing regularizers, like the Lasso \cite{tibshirani1996regression} or group lasso \cite{yuan2006model}, help scale linear Granger causality estimation in VAR models to the high-dimensional setting \cite{lozano2009groupedtemporal}, \cite{basu2015regularized}.

In classical linear VAR methods, one must explicitly specify the maximum time lag to consider when assessing Granger causality. If the specified lag is too short, Granger causal connections occurring at longer time lags between series will be missed while overfitting may occur if the lag is too large. Lag selection penalties, like the hierarchical lasso \cite{nicholson2014hierarchical} and truncating penalties \cite{shojaie2010discovering}, have been used to automatically select the relevant lags while protecting against overfitting. Furthermore, these penalties lead to a sparse network of Granger causal interactions, where only a few Granger causal connections exist for each series---a crucial property for scaling Granger causal estimation to the high-dimensional setting, where the number of time series and number of potentially relevant time lags all scale with the number of observations \cite{buhlmann2011statistics}.

Model-based methods may fail in real world cases when the relationships between the past of one series and future of another falls outside of the model class \cite{terasvirta2010modelling}, \cite{tong2011nonlinear}, \cite{lusch2016inferring}. This typically occurs when there are nonlinear dependencies between the past of one series and the future. Model-free methods, like transfer entropy \cite{vicente2011transfer} or directed information \cite{amblard2011directed}, are able to detect these nonlinear dependencies between past and future with minimal assumptions about the predictive relationships. However, these estimators have high variance and require large amounts of data for reliable estimation. These approaches also suffer from a curse of dimensionality \cite{runge2012escaping} when the number of series grows, making them inappropriate in the high-dimensional setting.

Neural networks are capable of representing complex, nonlinear, and non-additive interactions between inputs and outputs. Indeed, their time series variants, such as autoregressive multilayer perceptrons (MLPs) \cite{raissi2018multistep}, \cite{kicsi2004river}, \cite{billings2013nonlinear} and recurrent neural networks (RNNs) like long-short term memory networks (LSTMs) \cite{graves2012supervised} have shown impressive performance in forecasting multivariate time series given their past \cite{yu2017long}, \cite{zhang2003time}, \cite{li2017graph}. While these methods have shown impressive predictive performance, they are essentially black box methods and provide little interpretability of the multivariate structural relationships in the series. A second drawback is that jointly modeling a large number of series leads to many network parameters. As a result, these methods require much more data to fit reliably and tend to perform poorly in high-dimensional settings.

We present a framework for structure learning in MLPs and RNNs that leads to interpretable nonlinear Granger causality discovery. The proposed framework harnesses the impressive flexibility and representational power of neural networks. It also sidesteps the black-box nature of many network architectures by introducing component-wise architectures that disentangle the effects of lagged inputs on individual output series. For interpretability and an ability to handle limited data in the high-dimensional setting, we place sparsity-inducing penalties on particular groupings of the weights that relate the histories of individual series to the output series of interest. We term these sparse component-wise models, e.g. cMLP and cLSTM, when applied to the MLP and LSTM, respectively. In particular, we select for Granger causality by adding group sparsity penalties \cite{yuan2006model} on the outgoing weights of the inputs.

As in linear methods, appropriate lag selection is crucial for Granger causality selection in nonlinear approaches---especially in highly parametrized models like neural networks. For the MLP, we introduce two more structured group penalties \cite{nicholson2014hierarchical}, \cite{huang2011learning} \cite{kim2010tree} that automatically detect both nonlinear Granger causality and also the lags of each inferred interaction. Our proposed cLSTM model, on the other hand, sidesteps the lag selection problem entirely because the recurrent architecture efficiently models long range dependencies \cite{graves2012supervised}. When the true network of nonlinear interactions is sparse, both the cMLP and cLSTM approaches will select a subset of the time series that Granger-cause the output series, no matter the lag of interaction. To our knowledge, these approaches represent the first set of nonlinear Granger causality methods applicable in high dimensions without requiring precise lag specification.

We first validate our approach and the associated penalties via simulations on both linear VAR and nonlinear Lorenz-96 data \cite{karimi2010extensive}, showing that our nonparametric approach accurately selects the Granger causality graph in both linear and nonlinear settings. Second, we compare our cMLP and cLSTM models with existing Granger causality approaches \cite{lim2015operator}, \cite{lebre2009inferring} on the difficult DREAM3 gene regulatory network recovery benchmark datasets \cite{prill2010towards} and find that our methods outperform a wide set of competitors across all five datasets. Finally, we use our cLSTM method to explore Granger causal interactions between body parts during natural motion with a highly nonlinear and complex dataset of human motion capture \cite{cmu2009motion}, \cite{fox2014joint}. Our implementation is available online: \url{https://github.com/iancovert/Neural-GC}.

Traditionally, the success stories of neural networks have been on prediction tasks in large datasets. In contrast, here, our performance metrics relate to our ability to produce interpretable structures of interaction amongst the observed time series. Furthermore, these successes are achieved in limited data scenarios. Our ability to produce interpretable structures and train neural network models with limited data can be attributed to our use of structured sparsity-inducing penalties and the regularization such penalties provide, respectively. We note that sparsity inducing penalties have been used for architecture selection in neural networks \cite{alvarez2016learning}, \cite{louizos2017bayesian}. However, the focus of the architecture selection was on improving predictive performance rather than on returning interpretable structures of interaction among observed
quantities.

More generally, our proposed formulation shows how structured penalties common in regression \cite{huang2011learning}, \cite{kim2010tree} may be generalized for structured sparsity and regularization in neural networks. This opens up new opportunities to use these tools in other neural network context, especially as applied to structure learning problems. In concurrent work, a similar notion of sparse-input neural networks were developed for high-dimensional regression and classification tasks for independent data \cite{feng2017sparse}.

\section{Linear Granger Causality}\label{sec:linear_cause}
Let ${\bf x}_t  \in \mathbb{R}^p$ be a $p$-dimensional stationary time series and assume we have observed the process at $T$ time points, $({\bf x}_1, \ldots, {\bf x}_T)$. Using a model-based approach, as is our focus, Granger causality in time series analysis is typically studied using the vector autoregressive model (VAR) \cite{lutkepohl2005new}. In this model, the time series at time $t$, ${\bf x}_t$, is assumed to be a linear combination of the past $K$ lags of the series
\begin{equation}
{\bf x}_t = \sum_{k = 1}^K A^{(k)} {\bf x}_{t-k} + e_t,
\end{equation}
where $A^{(k)}$ is a $p \times p$ matrix that specifies how lag $k$ affects the future evolution of the series and $e_t$ is zero mean noise. In this model, time series $j$ does not Granger-cause time series $i$ if and only if for all $k$, $A_{ij}^{(k)} = 0$. A Granger causal analysis in a VAR model thus reduces to determining which values in $A^{(k)}$ are zero over all lags. In higher dimensional settings, this may be determined by solving a group lasso regression problem \cite{lozano2009grouped}
\begin{align} \label{eq:group_var}
\min_{A^{(1)}, \ldots, A^{(K)}} &\sum_{t = K}^T \| {\bf x}_t- \sum_{k = 1}^K A^{(k)} {\bf x}_{t - k}\|_2^2 \nonumber \\
&+ \lambda \sum_{ij} \|(A_{ij}^{(1)}, \ldots, A^{(K)}_{ij}\|_2,
\end{align}
where $\|\cdot\|_2$ denotes the $L_2$ norm. 
The group lasso penalty over all lags of each $(i,j)$ entry, $\|(A_{ij}^{(1)}, \ldots, A^{(K)}_{ij}\|_2$ jointly shrinks all $A_{ij}^k$ parameters to zero across all lags $k$ \cite{yuan2006model}. 
The hyper-parameter $\lambda > 0$ controls the level of group sparsity.

The group penalty in Equation (\ref{eq:group_var}) may be replaced with a structured hierarchical penalty \cite{huang2011learning}, \cite{jenatton2011proximal} that automatically selects the lag of each Granger causal interaction \cite{nicholson2014hierarchical}. Specifically, the hierarchical lag selection problem is given by 
\begin{align} \label{eq:h_var}
\min_{A^{(1)}, \ldots, A^{(K)}} &\sum_{t = K}^T \| {\bf x}_t - \sum_{k = 1}^K A^{(k)} {\bf x}_{t - k}\|_2^2 \nonumber \\
&+ \lambda  \sum_{ij} \sum_{k = 1}^K \|(A_{ij}^{(k)}, \ldots, A^{(K)}_{ij}\|_2,
\end{align}
where $\lambda > 0$ now controls the lag order selected for each
interaction. Specifically, at higher values of $\lambda$ there exists a $k$
for each $(i, j)$ pair such that the entire contiguous set of lags
$(A^{(k)}_{ij}, \ldots, A^{(K)}_{ij})$ is shrunk to zero. If $k = 1$ for a particular $(i,j)$ pair, then all lags are equal to zero and series $i$ does not Granger-cause series $j$; thus, this penalty simultaneously selects for Granger non-causality and the lag of each Granger causal pair.

\section{Models for Neural Granger Causality}
\subsection{Adapting Neural Networks for Granger Causality}
A \textit{nonlinear} autoregressive model (NAR) allows ${\bf x}_t$ to evolve according to more general nonlinear dynamics \cite{billings2013nonlinear}
\begin{align}
{\bf x_t} =g(x_{<t1} ,\ldots,x_{<tp} )+ e_t
\end{align}
where $x_{<t i} = \left(\ldots,  x_{(t - 2)i}, x_{(t - 1)i} \right)$ denotes the past of series $i$ and we assume additive zero mean noise $e_t$.

In a forecasting setting, it is common to jointly model the full nonlinear functions $g$ using neural networks. Neural networks have a long history in NAR forecasting, using both traditional architectures \cite{billings2013nonlinear}, \cite{chu1990neural}, \cite{billings1996determination} and more recent deep learning techniques \cite{yu2017long}, \cite{li2017graph}, \cite{tao2018hierarchical}. These approaches either utilize an MLP where the inputs are $x_{<t} = x_{(t-1):(t-K)}$, for some lag $K$, or a recurrent network, like an LSTM.

There are two problems with applying the standard neural network NAR model in the context of inferring Granger causality. The first is that these models act as black boxes that are difficult to interpret. Due to sharing of hidden layers, it is difficult to specify sufficient conditions on the weights that simultaneously allows series $j$ to Granger cause series $i$ but not Granger cause series $i'$ for $i \neq i'$. Second, a joint network over all $x_{ti}$ for all $i$ assumes that each time series depends on the same past lags of the other series. However, in practice, each $x_{ti}$ may depend on different past lags of the other series.

To tackle these challenges, we propose a structured neural network approach to modeling and estimation. First, instead of modeling $g$ jointly across all outputs $x_t$, as is standard in multivariate forecasting, we instead focus on each output \textit{component} with a separate model:
\begin{align*}
x_{t i}=g_{i}\left(x_{<t 1}, \ldots, x_{<t p}\right)+e_{t i}
\end{align*}
Here, $g_i$ is a function that specifies how the past $K$ lags are mapped to series $i$. In this context, Granger non-causality between two series $j$ and $i$ means that the function $g_i$ does not depend on $x_{<tj}$, the past lags of series $j$. More formally,

\newtheorem{definition}{Definition}
\begin{definition} \label{def:granger}
Time series $j$ is Granger non-causal for time series
$i$ if for all $\left(x_{<t 1}, \ldots, x_{<t p}\right)$ and all $x^{\prime}_{<tj}\neq x_{<t j}$,
\begin{align}
&g_{i}\left(x_{<t 1}, \ldots, x_{<t j}, \ldots, x_{<t p}\right) = \nonumber \\
&g_{i}\left(x_{<t 1}, \ldots, x^{\prime}_{<t j}, \ldots x_{<t p}\right) \nonumber
\end{align}
\noindent that is, $g_i$ is invariant to $x_{<tj}$.
\end{definition}

In Section \ref{subsec_sparseMLPs} and \ref{subsec_sparsecRNNs} we consider these component-wise models in the context of MLPs and LSTMs. We examine a set of sparsity inducing penalties as in Equations  (\ref{eq:group_var}) and (\ref{eq:h_var}) that allow us to infer the invariances of Definition \ref{def:granger} that lead us to identify Granger non-causality.

\begin{figure*}[ht]
\centering
\includegraphics[width=1.0\linewidth]{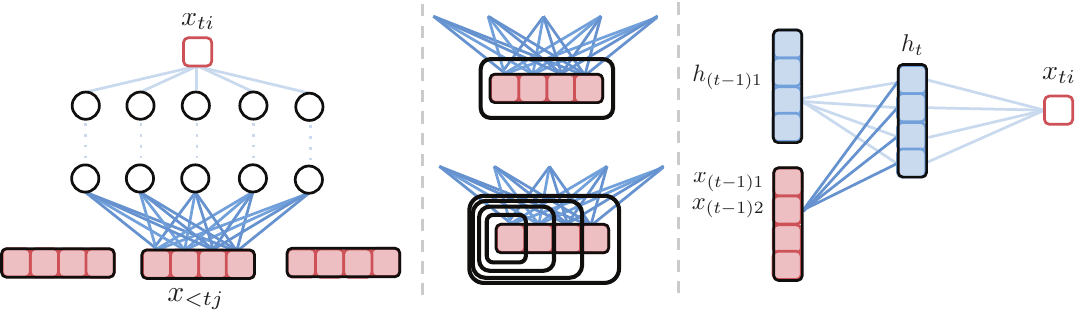}
\caption{(left) Schematic for modeling Granger causality using cMLPs. If the outgoing weights for series $j$, shown in dark blue, are penalized to zero, then series $j$ does not Granger-cause series $i$. (center) The group lasso penalty jointly penalizes the full set of outgoing weights while the hierarchical version penalizes the nested set of outgoing weights, penalizing higher lags more. (right) Schematic for modeling Granger causality using a cLSTM. If the dark blue outgoing weights to the hidden units from an input $x_{(t-1) j}$ are zero, then series $j$ does not Granger-cause series $i$.}
\label{Fig1}
\end{figure*}

\begin{figure}[ht]
\centering
\includegraphics[width=1.0\linewidth]{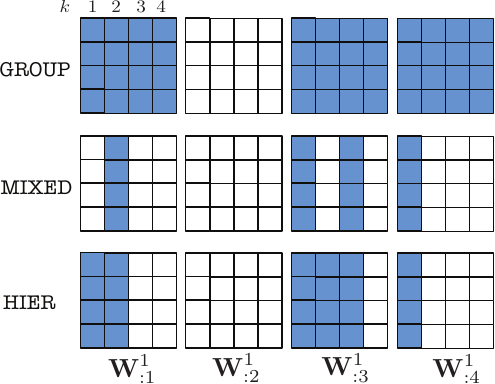}
\caption{Example of group sparsity patterns of the first layer weights of a
cMLP with four first layer hidden units and four input series with maximum
lag $k = 4$. Differing sparsity patterns are shown for the three different
structured penalties of group lasso (GROUP) from Equation (\ref{eq:group_pen}), group
sparse group lasso (MIXED) from Equation (\ref{h_pen}) and hierarchical lasso
(HIER) from Equation (\ref{eq:group_h_pen}).}
\label{Fig2}
\end{figure}

\begin{figure}[ht]
\centering
\includegraphics[width=1.0\linewidth]{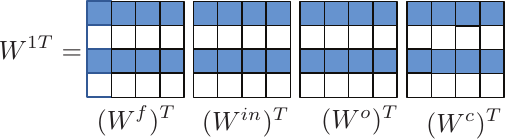}
\caption{Example of the group sparsity patterns in a sparse cLSTM model
with a four dimensional hidden state and four input series. Due to the
group lasso penalty on the columns of $W$, the $W^f$ , $W^{in}$, $W^o$, and $W^c$
matrices will share the same column sparsity pattern.}
\label{Fig3}
\end{figure}

\subsection{Sparse Input MLPs}\label{subsec_sparseMLPs}

Our first approach is to model each output component $g_i$ with a separate MLP, so that we can easily disentangle the effects from inputs to outputs. We refer to this approach, displayed pictorially in Figure \ref{Fig1}, as a \textit{component-wise} MLP (cMLP). Let $g_i$ take the form of an MLP with $L-1$ layers and let the vector $h_{t}^{l} \in \mathbb{R}^{H}$ denote the values of the $m$-dimensional $l$th hidden layer at time $t$. The parameters of the neural network are given by weights $\mathbf{W}$ and biases $\mathbf{b}$ at each layer, $\mathbf{W}=\left\{W^{1}, \ldots, W^{L}\right\}$ and $\mathbf{b}=\left\{b^{1}, \dots, b^{L}\right\}$. To draw an analogy with the time series VAR model, we further decompose the weights at the first layer across time lags, $W^{1}=\left\{W^{11}, \ldots, W^{1 K}\right\}$. The dimensions of the parameters are given by $W^{1} \in \mathbb{R}^{H \times p K}$, $W^{l} \in \mathbb{R}^{H \times H}$ for $1<l<L, W^{L} \in \mathbb{R}^{H}, b^{l} \in \mathbb{R}^{H}$ for $l<L$ and $b^{L} \in \mathbb{R}$. Using this notation, the vector of first layer hidden values at time $t$ are given by

\begin{equation}  \label{nn_layer}
h_{t}^{1}=\sigma\left(\sum_{k=1}^{K} W^{1 k} \mathbf{x}_{t-k}+b^{1}\right),
\end{equation} 
where $\sigma$ is an activation function. Typical activation functions are either \texttt{logistic} or \texttt{tanh} functions. The vector of hidden units in subsequent layers is given by a similar form, also with $\sigma$ activation functions:

\begin{equation}
h_{t}^{l}=\sigma\left(W^{l} h_{t}^{l-1}+b^{l}\right).
\end{equation}
After passing through the $L-1$ hidden layers, the time series output, $x_{ti}$, is given by a linear combination of the units in the final hidden layer
\begin{equation}
x_{t i}=g_{i}\left(x_{<t}\right)+e_{t i}=W^{L} h_{t}^{L-1}+b^{L}+e_{t i}
\end{equation}
where $W^L$ is the linear output decoder and $h_{t}^{L}$ is the final hidden output from the final $L-1$th layer. The error term, $e_{ti}$, is modeled as mean zero Gaussian noise. We chose this linear output decoder since our primary motivation involves real-valued multivariate time series. However, other decoders like a \texttt{logistic}, \texttt{softmax}, or poisson likelihood with exponential link function \cite{mccullagh1989generalized}, could be used to model nonlinear Granger causality in multivariate binary \cite{hall2016inference}, categorical \cite{tank2017granger}, or positive count time series \cite{hall2016inference}.

\subsubsection{Penalized Selection of Granger Causality in the cMLP}\label{subsubsec_penalizedcMLP}

In Equation (\ref{nn_layer}), if the $j$th column of the first layer weight matrix, $W_{ : j}^{1 k}$, contains zeros for all $k$, then series $j$ does not Granger-cause series $i$. That is, $x_{(t-k) j}$ for all $k$ does not influence the hidden unit $h_{t}^{1}$ and thus the output $x_{t i}$. Following Definition \ref{def:granger}, we see $g_i$ is invariant to $x_{<tj}$. Thus, analogously to the VAR case, one may select for Granger causality by applying a group penalty to the columns of the $W^{1 k}$ matrices for each $g_i$,
\begin{equation} \label{nn_pen}
\min _{\mathbf{W}} \sum_{t=K}^{T} \Big(x_{i t}-g_{i}\left(x_{(t-1) :(t-K)}\right) \Big)^2+\lambda \sum_{j=1}^{p} \Omega\left(W_{ : j}^{1}\right).
\end{equation}
where $\Omega$ is a penalty that shrinks the entire set of first layer weights for input series $j$, i.e., $W_{ : j}^{1}=\left(W_{ : j}^{11}, \ldots, W_{ : j}^{1 K}\right)$, to zero. We consider three different penalties that, together, show how we recast structured regression penalties to the
neural network case.

We first consider a group lasso penalty over the entire set of outgoing weights across all lags for time series $j$, $W_{ : j}^{1}$,

\begin{equation} \label{eq:group_pen}
\Omega\left(W_{ : j}^{1}\right)=\left\|W_{ : j}^{1}\right\|_{F},
\end{equation}
where $\|\cdot\|_{F}$ is the Froebenius matrix norm. This penalty shrinks all weights associated with lags for input series $j$ equally. For large enough $\lambda$, the solutions to Equation (\ref{nn_pen}) with the group penalty in Equation (\ref{eq:group_pen}) will lead to many zero columns in each $W^{1 k}$ matrix, implying only a small number of estimated Granger causal connections. This group penalty is the neural network analogue of the group lasso penalty across lags in Equation \ref{eq:group_var} for the VAR case.

To detect the lags where Granger causal effects exists, we propose a new penalty called a \textit{group sparse group lasso} penalty. This penalty assumes that only a few lags of a series $j$ are predictive of series $i,$ and provides both sparsity across groups (a sparse set of Granger causal time series) and sparsity within groups (a subset of relevant lags)
\begin{equation}\label{h_pen}
\Omega\left(W_{ : j}^{1}\right)=\alpha\left\|W_{ : j}^{1}\right\|_{F}+(1-\alpha) \sum_{k=1}^{K}\left\|W_{ : j}^{1 k}\right\|_{2}
\end{equation}
where $\alpha \in(0,1)$ controls the tradeoff in sparsity across and within groups. This penalty is a related to, and is a generalization of, the sparse group lasso \cite{doi:10.1080/10618600.2012.681250}.

Finally, we may simultaneously select for both Granger causality and the lag order of the interaction by replacing the group lasso penalty in Equation (\ref{nn_pen}) with a \textit{hierarchical} group lasso penalty \cite{nicholson2014hierarchical} in the MLP optimization problem,
\begin{equation} \label{eq:group_h_pen}
\Omega\left(W_{ : j}^{1}\right)=\sum_{k=1}^{K}\left\|\left(W_{ : j}^{1 k}, \ldots, W_{ : j}^{1 K}\right)\right\|_{F}.
\end{equation}
The hierarchical penalty leads to solutions such that for each $j$ there exists a lag $k$ such that all $W_{ : j}^{1 k^{\prime}}=0$ for $k^{\prime}>k$ and all $W_{ : j}^{1 k^{\prime}} \neq 0$ for $k^{\prime} \leq k$. Thus, this penalty effectively selects the lag of each interaction. The hierarchical penalty also sets many columns of $W^{1 k}$ to be zero across all $k$, effectively selecting for Granger causality. In practice, the hierarchical penalty allows us to fix $K$ to a large value, ensuring that no Granger causal connections at higher lags are missed. Example sparsity patterns selected by the three penalties are shown in Figure \ref{Fig2}.

While the primary motivation of our penalties is for efficient Granger causality selection, the lag selection penalties in Equations (\ref{h_pen}) and (\ref{eq:group_h_pen}) are also of independent interest to nonlinear forecasting with neural networks. In this case, over-specifying the lag of a NAR model leads to poor generalization and overfitting \cite{billings2013nonlinear}. One proposed technique in the literature is to first select the appropriate lags using forward orthogonal least squares \cite{billings2013nonlinear}; our approach instead combines model fitting and lag selection into one procedure.

\subsection{Sparse Input RNNs}\label{subsec_sparsecRNNs}
Recurrent neural networks (RNNs) are particularly well suited for modeling time series, as they compress the past of a time series into a hidden state, aiming to capture complicated nonlinear dependencies at longer time lags than traditional time series models. As with MLPs, time series forecasting with RNNs typically proceeds by jointly modeling the entire evolution of the multivariate series using a single recurrent network.

As in the MLP case, it is difficult to disentangle how each series affects the evolution of another series when using an RNN. This problem is even more severe in complicated recurrent networks like LSTMs. To model Granger causality with RNNs, we follow the same strategy as with MLPs and model each $g_i$ function using a separate RNN, which we refer to as a component-wise RNN (cRNN). For simplicity, we assume a single-layer RNN, but our formulation may be easily generalized to accommodate more layers.

Consider an RNN for predicting a single component. Let ${\bf h}_t \in \mathbb{R}^H$ represent the $H$-dimensional hidden state at time $t$, representing the historical context of the time series for predicting a component $x_{ti}$. The hidden state at time $t + 1$ is updated recursively 
\begin{align} \label{eq:hidden}
{\bf h}_t = f_i({\bf x}_t, {\bf h}_{t-1}),
\end{align}
where $f_i$ is some nonlinear function that depends on the particular recurrent architecture.

Due to their effectiveness at modeling complex time dependencies, we choose to model the recurrent function $f$ using an LSTM \cite{graves2012supervised}. The LSTM model introduces a second hidden state variable ${\bf c}_t$, referred to as the cell state, giving the full set of hidden parameter as $({\bf c}_t , {\bf h}_t )$. The LSTM model updates its hidden states recursively as
\begin{equation}
\label{eq:lstm}
\begin{aligned} 
{\bf f}_t &= \sigma \left(W^f {\bf x}_t + U^f {\bf h}_{(t - 1)} \right) \\
{\bf i}_t &= \sigma \left(W^{in} {\bf x}_t + U^{in} {\bf h}_{(t - 1)} \right) \\
{\bf o}_t &= \sigma \left(W^{o} {\bf x}_t + U^{o} {\bf h}_{(t - 1)} \right) \\
{\bf c}_t &= {\bf f}_t \odot {\bf c}_{t-1} + {\bf i}_t \odot \sigma \left(W^c {\bf x}_t + U^c {\bf h}_{t - 1} \right) \\
{\bf h}_t &= {\bf o}_t \odot \sigma ({\bf c}_t)
\end{aligned}
\end{equation}
where $\odot$ denotes element-wise multiplication and ${\bf i}_t$ , ${\bf f}_t$ , and ${\bf o}_t$ represent input, forget and output gates, respectively, that control how each component of the state cell, ${\bf c}_t$, is updated and then transferred to the hidden state used for prediction, ${\bf h}_t$. In particular, the forget gate, ${\bf f}_t$, controls the amount that the past cell state influences the future cell state, and the input gate, ${\bf i}_t$, controls the amount that the current observation influences the new cell state. The additive form of the cell state update in the LSTM allows it to encode long-range dependencies, since cell states from far in the past may still influence the cell state at time $t$ if the forget gates remain close to one. In the context of Granger causality, this flexible architecture can represent long-range, nonlinear dependencies between time series. As in the cMLP, the output for series $i$ at time $t$ is given by a linear decoding of the hidden state 
\begin{equation}\label{eqn:sparseinput}
x_{ti} = g_i(x_{<t}) + e_{ti} = W^2 {\bf h}_t + e_{ti},
\end{equation}
where $W^2$ are the output weights. We let ${\bf W} = (W^1, W^2, U^1)$ be the full set of parameters where $W^1 = \left( (W^f)^\top,(W^{in})^\top ,(W^o)^\top ,(W^c)^\top \right) ^\top$ and $U^1 = \left( (U^f)^\top,(U^{in})^\top ,(U^o)^\top ,(U^c)^\top  \right)^\top$ represent the full set of first layer weights. As in the MLP case, other decoding schemes could be used in the case of categorical or count data.

\subsubsection{Granger Causality Selection in LSTMs}
In Equation \eqref{eqn:sparseinput}  the set of input matrices $W^1$ controls how the past time series affect the forget gates, input gates, output gates, and cell updates, and, consequently, the update of the hidden representation. Like in the MLP case, for this component-wise LSTM model (cLSTM) a sufficient condition for Granger non-causality of an input series $j$ on an output $i$ is that all elements of the $j$th column of $W^1$ are zero, $W^1_{:j} = 0$. Thus, we may select series that Granger-cause series $i$ using a group lasso penalty across columns of $W^1$ by
\begin{align} \label{eq:nn_group_lasso}
\min_{\mathbf{W}} \sum_{t = 2}^T \Big(x_{it} - g_{i}(x_{<t}) \Big)^2 + \lambda \sum_{j = 1}^p ||W^1_{:j}||_2.
\end{align}
For a large enough $\lambda$, many columns of $W^1$ will be zero, leading to a sparse set of Granger causal connections. An example sparsity pattern in the  LSTM parameters is shown in Figure \ref{Fig3}.

\section{Optimizing the Penalized Objectives}

\subsection{Optimizing the Penalized cMLP Objective}
We optimize the nonconvex objectives of Equation (\ref{nn_pen}) using proximal gradient descent \cite{parikh2014proximal}. Proximal optimization is important in our context because it leads to exact zeros in the columns of the input matrices, a critical requirement for interpretating Granger non-causality in our framework. Additionally, a line search can be incorporated into the optimization algorithm to ensure convergence to a local minimum \cite{gong2013general}.
The algorithm updates the network weights ${\bf W}$ iteratively starting with ${\bf W}^{(0)}$ by
\begin{equation}
{\bf W}^{(m + 1)} = \text{prox}_{\gamma^{(m)} \lambda \Omega} \left( {\bf W}^{(m)} - \gamma^{(m)} \nabla \mathcal{L} ({\bf W}^{(m)}) \right),
\end{equation}
where $\mathcal{L} = \sum_{t = K}^{T} \big(x_{ti} - g_i(x_{<t})\big)^2$ is the the neural network prediction loss and $\text{prox}_{\lambda \Omega}$ is the proximal operator with respect to the sparsity inducing penalty function $\Omega$. The entries in ${\bf W}^{(0)}$ are initialized randomly from a standard normal distribution. The scalar $\gamma^{(m)}$ is the step size, which is either set to a fixed value or determined by line search \cite{gong2013general}.
While the objectives in Equation (\ref{nn_pen}) are nonconvex, we find that no random restarts are required to accurately detect Granger causality connections. 

Since the sparsity promoting group penalties are only applied to the input weights, the proximal step for weights at the higher levels is simply the identity function. The proximal step for the group lasso penalty on the input weights is given by a group soft-thresholding operation on the input weights \cite{parikh2014proximal},

\begin{align}
\text{prox}_{\gamma^{(m)} \lambda \Omega}(W^i_{:k}) &= \text{soft} (W_{:k}^1, \gamma^{(m)} \lambda) \label{proxgrouplasso}\\
&= \left(1 - \frac{\lambda \gamma^{(m)}}{||W^1_{:j}||_{F}} \right)_{+} W^1_{:k},
\end{align}
where $(x)_+ = \max(0, x)$. For the group sparse group lasso, the proximal step on the input weights is given by group-soft thresholding on the lag specific weights, followed by group soft thresholding on the entire resulting input weights for each series, see Algorithm \ref{alg:prox_group}. The proximal step on the input weights for the hierarchical penalty is given by iteratively applying the group soft-thresholding operation on each nested group in the penalty, from the smallest group to the largest group \cite{jenatton2011proximal}, and is shown in Algorithm \ref{alg:prox_h}.

\begin{algorithm}[H]
\caption{Proximal gradient descent with line search
algorithm for solving Equation (\ref{nn_pen}).
Proximal steps given in Equation (\ref{proxgrouplasso}) for the group lasso penalty, in Algorithm~\ref{alg:prox_group} for the group sparse group lasso penalty, and in Algorithm~\ref{alg:prox_h} for the hierarchical penalty.
}
\label{algorithm1}
\begin{algorithmic}
\REQUIRE{$\lambda>0$}
\STATE{$m=0$, initialize $\mathbf{W}^{(0)}$ }
\WHILE{not converged}
\STATE{$m=m+1$}
\STATE{determine $\gamma$ by line search}
\FOR{$j=1$ to $p$}
\STATE{$W_{ : j}^{1(m+1)}=\operatorname{prox}_{\gamma \lambda \Omega}\left(W_{ : j}^{1(m)} - \gamma \nabla_{W_{ : j}^{1}} \mathcal{L}\left(\mathbf{W}^{(m)}\right)\right)$}
\ENDFOR{}
\FOR{$l=2$ to $L$}
\STATE{$W^{l(m+1)}=W^{l(m)} - \gamma \nabla_{W^{l}} \mathcal{L}\left(\mathbf{W}^{(m)}\right)$}
\ENDFOR{}
\ENDWHILE{}
\RETURN{($\mathbf{W}^{(m)}$)}
\end{algorithmic}
\end{algorithm}

\begin{algorithm}[H]
\caption{One pass algorithm to compute the proximal map for the group sparse group lasso penalty, for relevant lag selection in the cMLP model.}
 \label{alg:prox_group}
\begin{algorithmic}
\REQUIRE{$\lambda>0, \gamma>0,\left(W_{ : j}^{11}, \ldots, W_{ : j}^{1 K}\right)$}
\FOR{$k=K$ to $1$}
\STATE{$W_{ : j}^{1 k}=\operatorname{soft}\left(W_{ : j}^{1 k}, \gamma \lambda\right)$}
\ENDFOR{}
\STATE{$\left(W_{ : j}^{11}, \ldots, W_{ : j}^{1 K}\right)=\operatorname{soft}\left(\left(W_{ : j}^{11}, \ldots, W_{ : j}^{1 K}\right), \gamma \lambda\right)$}
\RETURN{$\left(W_{ : j}^{11}, \ldots, W_{ : j}^{1 K}\right)$}
\end{algorithmic}
\end{algorithm}
\begin{algorithm}[H] 
\caption{One pass algorithm to compute the proximal
map for the hierarchical group lasso penalty, for automatic
lag selection in the cMLP model.}
\label{alg:prox_h}
\begin{algorithmic}
\REQUIRE{$\lambda>0, \gamma>0,\left(W_{ : j}^{11}, \ldots, W_{ : j}^{1 K}\right)$}
\FOR{$k=K$ to $1$}
\STATE{$\left(W_{ : j}^{1 k}, \ldots, W_{ : j}^{1 K}\right)=\operatorname{soft}\left(\left(W_{ : j}^{1 k}, \ldots, W_{ : j}^{1 K}\right), \gamma \lambda\right)$}
\ENDFOR{}
\RETURN{$\left(W_{ : j}^{11}, \ldots, W_{ : j}^{1 K}\right)$}
\end{algorithmic}
\end{algorithm}

Since all datasets we study are relatively small, the gradients are with respect to the full data objective (i.e., all time points); for larger datasets, one could instead use proximal stochastic gradient descent \cite{xiao2014proximal}.

\subsection{Optimizing the Penalized cLSTM Objective}
Similar to the cMLP, we optimize Equation (\ref{eq:nn_group_lasso}) using proximal gradient descent. When the data consists of many replicates of short time series, like in the DREAM3 data in Section \ref{sec_DREAM}, we perform a full backpropagation through time (BPTT) to compute the gradients. However, for longer series we truncate the BPTT by unlinking the hidden sequences. In practice, we do this by splitting the dataset up into equal sized batches, and treating each batch as an independent realization. Under this approach, the gradients used to optimize Equation (\ref{eq:nn_group_lasso}) are only approximations of the gradients of the full cLSTM model. This is very common practice in the training of of RNNs \cite{williams1995gradient}, \cite{sutskever2013training}, \cite{werbos1990backpropagation}. The full optimization algorithm for training is shown in Algorithm \ref{algorithm4}.

\begin{algorithm}[H]
\caption{Proximal gradient descent with line search algorithm for solving Equation (\ref{eq:nn_group_lasso}) for the cLSTM with group lasso penalty.}
\label{algorithm4}
\begin{algorithmic}
\REQUIRE{$\lambda>0$}
\STATE{$m=0$, initialize $\mathbf{W}^{(0)}$ }
\WHILE{not converged}
\STATE{$m=m+1$}
\STATE{compute $\nabla \mathcal{L}\left(\mathbf{W}^{(m)}\right)$ by BPTT (truncated for large $T$)}
\STATE{determine $\gamma$ by line search.}
\FOR{$j=1$ to $p$}
\STATE{$W_{ : j}^{1(m+1)}=\operatorname{soft}\left(W_{ : j}^{1(m)} - \gamma \nabla_{W_{ : j}^{1}} \mathcal{L}\left(\mathbf{W}^{(m)}\right), \gamma \lambda\right)$}
\ENDFOR{}
\STATE{$W^{2(m+1)}=W^{2(m)} - \gamma \nabla_{W^{2}} \mathcal{L}\left(\mathbf{W}^{(m)}\right)$}
\STATE{$U^{1(m+1)}=U^{1(m)} - \gamma \nabla_{U^1} \mathcal{L}\left(\mathbf{W}^{(m)}\right)$}
\ENDWHILE{}
\RETURN{($\mathbf{W}^{(m)}$)}
\end{algorithmic}
\end{algorithm}


\begin{table*}[!t]
  \caption{Comparison of AUROC for Granger causality selection among different approaches,
  as a function of the forcing constant $F$ and the length of the time series $T$. Results are the mean across five initializations, with 95\% confidence intervals.}
  \label{Table1}
  \centering 
  \begin{tabular}{lcccccc}\toprule
    Model & \multicolumn{3}{c}{$F$ = 10} & \multicolumn{3}{c}{$F$ = 40} \\
    \cmidrule(r){1-1} \cmidrule(r){2-4} \cmidrule(r){5-7}
    $T$ & 250 & 500 & 1000 & 250 & 500 & 1000 \\ \midrule
    cMLP & \textbf{86.6 $\pm$ 0.2} & \textbf{96.6 $\pm$ 0.2} & \textbf{98.4 $\pm$ 0.1} & \textbf{84.0 $\pm$ 0.5} & \textbf{89.6 $\pm$ 0.2} & \textbf{95.5 $\pm$ 0.3} \\
    cLSTM & 81.3 $\pm$ 0.9 & 93.4 $\pm$ 0.7 & 96.0 $\pm$ 0.1 & 75.1 $\pm$ 0.9 & 87.8 $\pm$ 0.4 & 94.4 $\pm$ 0.5 \\
    IMV-LSTM & 63.7 $\pm$ 4.3 & 76.0 $\pm$ 4.5 & 85.5 $\pm$ 3.4 & 53.6 $\pm$ 5.2 & 59.0 $\pm$ 4.5 & 69.0 $\pm$ 4.8 \\
    LOO-LSTM & 47.9 $\pm$ 3.2 & 49.4 $\pm$ 1.8 & 50.1 $\pm$ 1.0 & 50.1 $\pm$ 3.3 & 49.1 $\pm$ 3.2 & 51.1 $\pm$ 3.7 \\
    \bottomrule
  \end{tabular}
\end{table*}

\section{Comparing cMLP and cLSTM Models for Granger Causality}
Both the cMLP and cLSTM frameworks model each component function $g_i$ using independent networks for each $i$. For the cMLP model, one needs to specify a maximum possible model lag $K$. However, our lag selection strategy (Equation~\ref{eq:group_h_pen}) allows one to set $K$ to a large value and the weights for higher lags are automatically removed from the model. On the other hand, the cLSTM model requires no maximum lag specification, and instead automatically learns the memory of each interaction. As a consequence, the cMLP and cLSTM differ in the amount of data used for training, as noted by a comparison of the $t$ index in Equation (\ref{eq:nn_group_lasso}) and Equation~(\ref{eq:group_h_pen}). For a length $T$ series, the cMLP and cLSTM models use $T - K$ and $T - 1$ data points, respectively. While insignificant for large $T$ , when the data consist of independent replicates of short series, as in the DREAM3 data in Section \ref{sec_DREAM}, the difference may be important. This ability to simultaneously model longer range dependencies while harnessing the full training set may explain the impressive performance of the cLSTM in the DREAM3 data in Section \ref{sec_DREAM}.

Finally, the zero outgoing weights in both the cMLP and cLSTM are a sufficient but not necessary condition to represent Granger non-causality. Indeed, series $i$ could be Granger non-causal of series $j$ through a complex configuration of weights that exactly cancel each other. However, because we wish to interpret the outgoing weights of the inputs as a measure of dependence, it is important that these weights reflect the true relationship between inputs and outputs. Our penalization schemes in both cMLP and cLSTM acts as a prior that biases the network to represent Granger non-causal relationships with zeros in the outgoing weights of the inputs, rather than through other configurations. Our simulation results in Section \ref{sec_simulation_expts} validate this intuition.

\section{Simulation Experiments}\label{sec_simulation_expts}

\subsection{cMLP and cLSTM Simulation Comparison}\label{subsec_cMLPcLSTMCompare}

To compare and analyze the performance of our two approaches, the cMLP and cLSTM, we apply both methods to detecting Granger causality networks in simulated linear VAR data and simulated Lorenz-96 data \cite{karimi2010extensive}, a nonlinear model of climate dynamics. Overall, the results show that our methods can accurately reconstruct the underlying Granger causality graph in both linear and nonlinear settings. We first describe the results from the Lorenz experiment and present the VAR results subsequently.

\subsubsection{Lorenz-96 Model}

\begin{figure}
\centering
\includegraphics[width=1.0\linewidth]{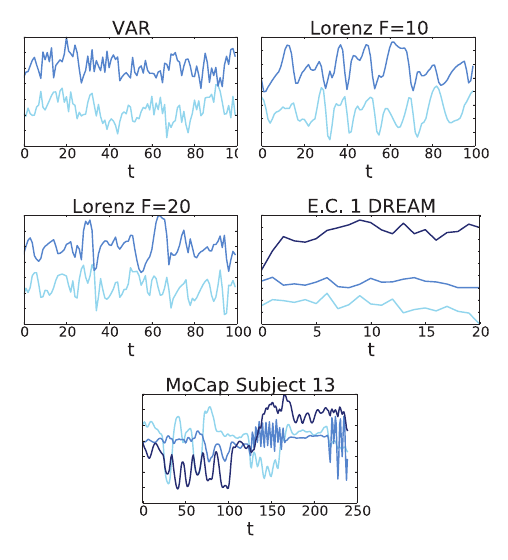}
\caption{Example multivariate linear (VAR) and nonlinear (Lorenz, DREAM,
and MoCap) series that we analyze using both cMLP and cLSTM models.
Note as the forcing constant, $F$, in the Lorenz model increases, the data
become more chaotic.}
\label{Fig4}
\end{figure}

The continuous dynamics in a $p$-dimensional Lorenz model are given by

\begin{equation}
\frac{d x_{t i}}{d t}=\left(x_{t(i+1)}-x_{t(i-2)}\right) x_{t(i-1)}-x_{t i}+F ,
\end{equation}
where $x_{t(-1)}=x_{t(p-1)}, x_{t 0}=x_{t p}, x_{t(p+1)}=x_{t 1}$ and $F$ is a
forcing constant that determines the level of nonlinearity and chaos in the series. Example series for two settings of $F$ are displayed in Figure \ref{Fig4}. We numerically simulate a $p = 20$ Lorenz-96 model with a sampling rate of $\Delta_{t}=0.05$, which results in a multivariate, nonlinear time series with sparse Granger causal connections.

\begin{table*}[!t]
  \caption{Comparison of AUROC for Granger causality selection among different approaches,
  as a function of the VAR lag order and the length of the time series $T$. Results are the mean across five initializations, with 95\% confidence intervals.}
  \label{Table2}
  \centering
  \begin{tabular}{lcccccc}\toprule
    Model & \multicolumn{3}{c}{VAR(1)} & \multicolumn{3}{c}{VAR(2)} \\
    \cmidrule(r){1-1} \cmidrule(r){2-4} \cmidrule(r){5-7}
    $T$ & 250 & 500 & 1000 & 250 & 500 & 1000 \\ \midrule
    cMLP & \textbf{91.6 $\pm$ 0.4} & \textbf{94.9 $\pm$ 0.2} & \textbf{98.4 $\pm$ 0.1} & \textbf{84.4 $\pm$ 0.2} & 88.3 $\pm$ 0.4 & 95.1 $\pm$ 0.2 \\
    cLSTM & 88.5 $\pm$ 0.9 & \textbf{93.4 $\pm$ 1.9} & 97.6 $\pm$ 0.4 & 83.5 $\pm$ 0.3 & \textbf{92.5 $\pm$ 0.9} & \textbf{97.8 $\pm$ 0.1} \\
    IMV-LSTM & 53.7 $\pm$ 7.9 & 63.2 $\pm$ 8.0 & 60.4 $\pm$ 8.3 & 53.5 $\pm$ 3.9 & 54.3 $\pm$ 3.6 & 55.0 $\pm$ 3.4 \\
    LOO-LSTM & 50.1 $\pm$ 2.7 & 50.2 $\pm$ 2.6 & 50.5 $\pm$ 1.9 & 50.1 $\pm$ 1.4 & 50.4 $\pm$ 1.4 & 50.0 $\pm$ 1.0 \\
    \bottomrule
  \end{tabular}
\end{table*}

Using this simulation setup, we test our models' ability to recover the underlying causal structure.
Average values of area under the ROC curve (AUROC) for recovery of the causal structure across five initialization seeds are shown in 
Table~\ref{Table1}, and we obtain results
under three different data set lengths, $T \in(250,500,1000)$, and two forcing constants, $F \in(10,40)$.

We compare results for the cMLP and cLSTM with two baseline methods that also rely on neural networks, the IMV-LSTM and leave-one-out LSTM (LOO-LSTM) approaches.
The IMV-LSTM \cite{guo2018interpretable}, \cite{guo2019exploring} uses attention weights to provide greater interpretability than standard LSTMs, and it detects Granger causal relationships by aggregating its attention weights. LOO-LSTM detects Granger causal relationships through the increase in loss that results from withholding each input time series (see Appendix).

We use $H = 100$ hidden units for all four methods, as experiments show that performance does not improve with a different number of units. While more layers may prove beneficial, for all experiments we fix the number of hidden layers, $L$, to one and leave the effects of additional hidden layers to future work. For the cMLP, we use the hierarchical penalty with model lag of $K = 5$; see Section \ref{subsec_quantanalysispenalty} for a performance comparison of several possible
penalties across model input lags.

For our methods, we compute AUROC values by sweeping $\lambda$ across a range of values; discarded edges (inferred Granger non-causality) for a particular $\lambda$ setting are those whose associated $L_2$ norm of the input weights of the neural network is equal to zero. Note that our proximal gradient algorithm sets many of these groups to be exactly zero. We compute AUROC values for the IMV-LSTM and LOO-LSTM by sweeping a range of thresholds for either the attention values or the increase in loss due to withholding time series.

As expected, the results indicate that the cMLP and
cLSTM performance improves as the data set size $T$ increases. The cMLP outperforms the cLSTM both in the less chaotic regime of $F = 10$ and the more chaotic regime of $F = 40$, but the gap in their performance narrows as more data is used. Both methods outperform the IMV-LSTM and LOO-LSTM by a wide margin.
Our models'
95\% confidence intervals are also relatively narrow, at less than 1\% AUROC for the cLSTM and cMLP, compared with 3-5\% for the IMV-LSTM.


To understand the role of the number of hidden units in our methods, we perform an ablation study to test different values of $H$; the results show that both the cMLP and cLSTM are robust to smaller $H$ values, but that their performance benefits from $H = 100$ hidden units (see Appendix). Additionally, we investigate the importance of the optimization algorithm; we found that Adam \cite{kingma2014adam}, proximal gradient descent \cite{parikh2014proximal} and proximal gradient descent with a line search \cite{gong2013general} lead to similar results (see Appendix). However, because Adam requires a thresholding parameter and the line search is computationally costly, we use standard proximal gradient descent in the remainder of our experiments.

\subsubsection{VAR Model}

To analyze our methods' performance when the true underlying dynamics are linear, we simulate data from $p = 20$ VAR(1) and VAR(2) models with randomly generated sparse transition matrices. To generate sparse dependencies for each time series $i$, we create self dependencies and randomly select three more dependencies among the other $p-1$ time series. Where series $i$ depends on series $j$, we set $A_{i j}^{k}=0.1$ for $k = 1$ or $k = 1,2$, and all other entries of $A$ are set to zero. Examining both VAR models allows us to see how well our methods detect Granger causality at longer time lags, even though no time lag is explicitly specified in our models.
Our results are the average over five random initializations for a single
dependency graph.

The AUROC results are displayed in Table~\ref{Table2} for the cLSTM, cMLP, IMV-LSTM, and LOO-LSTM approaches for three dataset lengths, $T \in(250,500,1000)$. The performance of the cLSTM and cMLP improves at larger $T$, and, as in the Lorenz-96 case, both models outperform the IMV-LSTM and LOO-LSTM by a wide margin. The cMLP remains more robust than the cLSTM with smaller amounts of data, but the cLSTM outperforms the cMLP on several occasions with $T = 500$ or $T = 1000$.

The IMV-LSTM consistently underperforms our methods with these datasets, likely because it is not explicitly designed for Granger causality discovery. Our finding that the IMV-LSTM performs poorly at this task is consistent with recent work suggesting that attention mechanisms are not indicative of feature importance \cite{wiegreffe2019attention}. The LOO-LSTM approach consistently achieves poor performance, likely due to two factors: (i)~unregularized LSTMs are prone to overfitting in the low-data regime, even when Granger causal time series are held out, and (ii)~withholding a single time series will not impact the loss if the remaining time series have dependencies that retain its signal.


\begin{table}[!t]
  \caption{AUROC comparisons between different cMLP Granger causality selection penalties on simulated Lorenz-96 data as a function of the input model lag, $K$. Results are the mean across five initializations, with 95\% confidence intervals.}
  \label{Table3}
  \centering
  \begin{tabular}{lccc}\toprule
    Lag $K$ & 5 & 10 & 20 \\
    \midrule
    GROUP & 88.1 $\pm$ 0.8 & 82.5 $\pm$ 0.3 & 80.5 $\pm$ 0.5 \\
    MIXED & 90.1 $\pm$ 0.5 & 85.4 $\pm$ 0.3 & 83.3 $\pm$ 1.1 \\
    HIER & \textbf{95.5 $\pm$ 0.2} & \textbf{95.4 $\pm$ 0.5} & \textbf{95.2 $\pm$ 0.3} \\
    \bottomrule
  \end{tabular}
\end{table}

\subsection{Quantitative Analysis of the Hierarchical Penalty}\label{subsec_quantanalysispenalty}

We next quantitatively compare the three possible structured penalties for Granger causality selection in the cMLP model. In Section \ref{subsec_sparseMLPs} we introduced the full group lasso (GROUP) penalty over all lags (Equation \ref{nn_pen}), the group sparse group lasso (MIXED) (Equation \ref{h_pen}) and the hierarchical (HIER) lag selection penalty (Equation \ref{eq:group_h_pen}). We compare these approaches across various choices of the cMLP model's maximum lag, $K \in(5,10,20)$. We use $H = 10$ hidden units for data simulated from the nonlinear Lorenz model with $F = 20$, $p = 20$, and $T = 750$. As in Section \ref{subsec_cMLPcLSTMCompare}, we compute the mean AUROC over five random initializations
and display the results in Table~\ref{Table3}. Importantly, the hierarchical penalty outperforms both group and mixed penalties across all model input lags $K$. Furthermore, performance significantly declines as $K$ increases in both group and mixed settings while the performance of the hierarchical penalty stays roughly constant as $K$ increases. This result suggests that performance of the hierarchical penalty for nonlinear Granger causality selection is robust to the input lag, implying that precise lag specification is unnecessary. In practice, this allows one to set the model lag to a large value without worrying that nonlinear Granger causality detection will be compromised.

\subsection{Qualitative Analysis of the Hierarchical Penalty}\label{subsec_analysispenalty}

To qualitatively validate the performance of the hierarchical group lasso penalty for automatic lag selection, we apply our penalized cMLP framework to data generated from a sparse VAR model with longer interactions. Specifically, we generate data from a $p = 10$, VAR(3) model as in Section \ref{sec:linear_cause}. To generate sparse dependencies for each time series $i$, we create self dependencies and randomly select two more dependencies among the other $p-1$ time series.  When series $i$ depends on series $j$, we set $A_{i j}^{k}=0.1$ for $k = 1,2,3$. All other entries of $A$ are set to zero. This implies that the Granger causal connections that do exist are of true lag 3. We run the cMLP with the hierarchical group lasso penalty and a maximal lag order of $K = 5$; for comparison, we also train a VAR model with a hierarchical penalty and maximal lag order $K = 5$.

\begin{figure}[ht]
\centering
\includegraphics[width=1.0\linewidth]{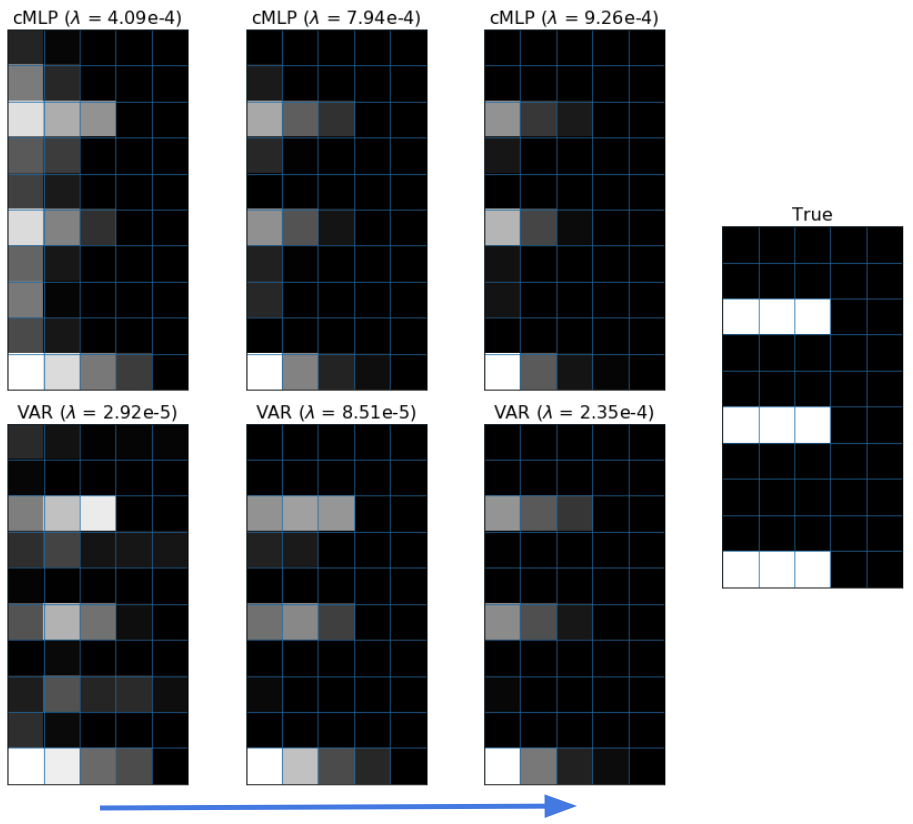}
\caption{Qualitative results of the cMLP automatic lag selection using a
hierarchical group lasso penalty and maximal lag of $K = 5$. The true
data are from a VAR(3) model. The images display results for a single
cMLP (one output series) and a VAR model using various penalty strengths $\lambda$. The rows
of each image correspond to different input series while the columns
correspond to the lag, with $k = 1$ at the left and $k = 5$ at the right. The
magnitude of each entry is the $L_2$ norm of the associated input weights
of the neural network after training. The true lag interactions are shown
in the rightmost image. Brighter color represents larger magnitude.
}
\label{Fig5}
\end{figure}

We visually display the selection results for one cMLP (i.e., one output series) and the VAR baseline across a variety of $\lambda$ settings in Figure \ref{Fig5}. For the lower $\lambda= 4.09e{-}4$ setting, the cMLP both (i) overestimates the lag order for a few input series and (ii) allows some false positive Granger causal connections. For the higher $\lambda= 7.94e{-}4$, lag selection performs almost perfectly, in addition to correct estimation of the Granger causality graph. Higher $\lambda$ values lead to larger penalization on longer lags, resulting in weaker long-lag connections. The VAR model, which is ideal for VAR data, does not perform noticeably better. While we show results for multiple $\lambda$ values for visualization, in practice one may use cross validation to select the appropriate $\lambda$.

\section{DREAM Challenge}\label{sec_DREAM}

We next apply our methods to estimate Granger causality
networks from a realistically simulated time course gene
expression data set. The data are from the DREAM3 challenge
\cite{prill2010towards} and provide a difficult, nonlinear data set for rigorously comparing Granger causality detection methods \cite{lim2015operator}, \cite{lebre2009inferring}.
The data is simulated using continuous gene expression and
regulation dynamics, with multiple hidden factors that are
not observed. The challenge contains five different simulated
data sets, each with different ground truth Granger causality
graphs: two E. Coli (E.C.) data sets and three Yeast (Y.) data
sets. Each data set contains $p = 100$ different time series, each
with $46$ replicates sampled at $21$ time points for a total of
$966$ time points. This represents a very limited data scenario
relative to the dimensionality of the networks and complexity
of the underlying dynamics of interaction. Three time series
components from a single replicate of the E. Coli 1 data set
are shown in Figure 4.

We apply both the cMLP and cLSTM to all five data sets.
Due to the short length of the series replicates, we choose
the maximum lag in the cMLP to be $K = 2$ and use $H = 10$ and
$H = 5$ hidden units for the cMLP and cLSTM, respectively. For
our performance metric, we consider the DREAM3 challenge
metrics of area under the ROC curve (AUROC) and area
under the precision recall curve (AUPR). Both curves are
computed by sweeping $\lambda$ over a range of values, as described
in Section \ref{sec_simulation_expts}.

\begin{figure}[ht]
\centering
\includegraphics[width=1.0\linewidth]{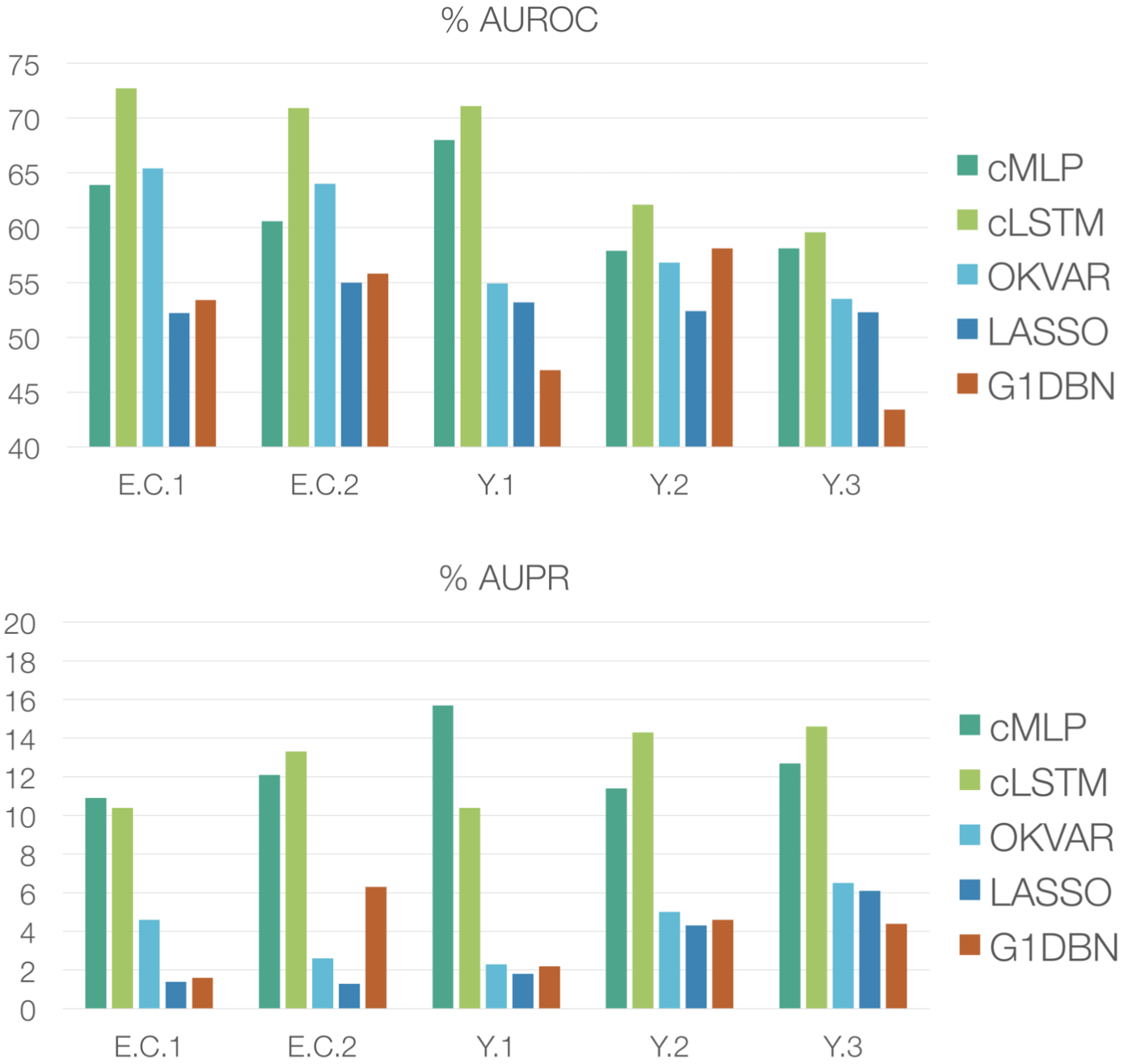}
\caption{(Top) AUROC and (bottom) AUPR (given in $\%$) results for our
proposed regularized cMLP and cLSTM models and the set of methods—
OKVAR, LASSO, and G1DBN presented in \cite{lim2015operator}. These results are for
the DREAM3 size-$100$ networks using the original DREAM3 data sets.}
\label{Fig6}
\end{figure}

\begin{figure}[ht]
\centering
\includegraphics[width=1.0\linewidth]{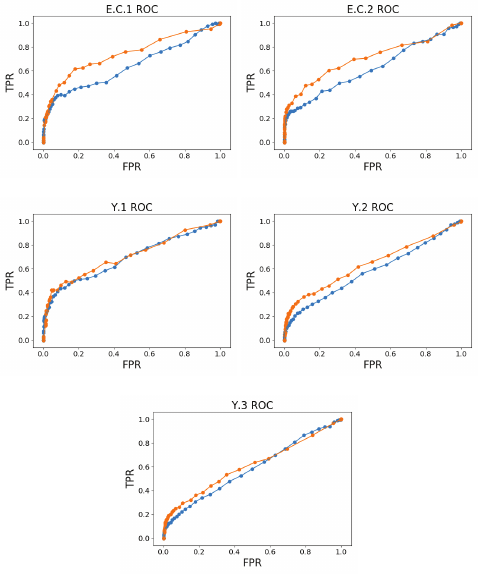}
\caption{ROC curves for the cMLP (\protect\blueline) and cLSTM (\protect\orangeline) models on the five
DREAM datasets.}
\label{Fig7}
\end{figure}

In Figure \ref{Fig6}, we compare the AUROC and AUPR of our
cMLP and cLSTM to previously published AUROC and
AUPR results on the DREAM3 data \cite{lim2015operator}. These comparisons
include both linear and nonlinear approaches: (i)~a linear
VAR model with a lasso penalty (LASSO) \cite{lozano2009groupedtemporal}, (ii)~a dynamic
Bayesian network using first-order conditional dependencies
(G1DBN) \cite{lebre2009inferring}, and (iii)~a state-of-the-art multi-output kernel
regression method (OKVAR) \cite{lim2015operator}. The latter is the most
mature of a sequence of nonlinear kernel Granger causality
detection methods \cite{sindhwani2012scalable}, \cite{marinazzo2008kernel}. In terms of AUROC, our cLSTM
outperforms all methods across all five datasets. Furthermore,
the cMLP method outperforms previous methods on two
datasets, Y.1 and Y.3, ties G1DBN on Y.2, and slightly under
performs OKVAR in E.C.1 and E.C.2. In terms of AUPR, both
cLSTM and cMLP methods do much better than all previous
approaches, with the cLSTM outperforming the cMLP in
three datasets. The raw ROC curves for cMLP and cLSTM
are displayed in Figure \ref{Fig7}.

These results clearly demonstrate the importance of
taking a nonlinear approach to Granger causality detection
in a (simulated) real-world scenario. Among the nonlinear
approaches, the neural network methods are extremely powerful.
Furthermore, the cLSTM’s ability to efficiently capture
long memory (without relying on long-lag specifications)
appears to be particularly useful. This result validates many
findings in the literature where LSTMs outperform autoregressive MLPs.
An interesting facet of these results, however, is that the
impressive performance gains are achieved in a limited
data scenario and on a task where the goal is recovery of
interpretable structure. This is in contrast to the standard
story of prediction on large datasets. To achieve these results,
the regularization and induced sparsity of our penalties is
critical.

\section{Dependencies in Human Motion Capture Data}

We next apply our methodology to detect complex, nonlinear
dependencies in human motion capture (MoCap) recordings.
In contrast to the DREAM3 challenge results, this analysis
allows us to more easily visualize and interpret the learned
network. Human motion has been previously modeled using
both linear dynamical systems \cite{Hsu:2005:STH:1073204.1073315}, switching linear dynamical
systems \cite{fox2014joint}, \cite{pavlovic2001learning} and also nonlinear dynamical models
using Gaussian processes \cite{wang2007gaussian}. While the focus of previous
work has been on motion classification \cite{Hsu:2005:STH:1073204.1073315} and segmentation
\cite{fox2014joint}, our analysis delves into the potentially long-range,
nonlinear dependencies between different regions of the
body during natural motion behavior. We consider a data
set from the CMU MoCap database \cite{cmu2009motion} previously studied
in \cite{fox2014joint}. The data set consists of $p = 54$ joint angle and body
position recordings across two different subjects for a total of
$T = 2024$ time points. In total, there are recordings from $24$
unique regions because some regions, like the thorax, contain
multiple angles of motion corresponding to the degrees of
freedom of that part of the body.

\begin{figure}[ht]
\centering
\includegraphics[width=1.0\linewidth]{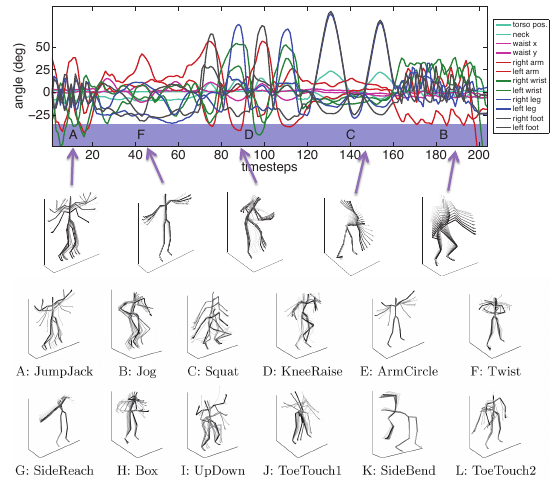}
\caption{(Top) Example time series from the MoCap data set paired with
their particular motion behaviors. (Bottom) Skeleton visualizations of $12$
possible exercise behavior types observed across all sequences analyzed
in the main text.}
\label{Fig8}
\end{figure}

We apply the cLSTM model with $H = 8$ hidden units
to this data set. For computational speed ups, we break the
original series into length $20$ segments and fit the penalized
cLSTM model from Equation (\ref{eq:nn_group_lasso}) over a range of $\lambda$ values. To
develop a weighted graph for visualization, we let the edge
weight $w_{ij}$ between components be the norm of the outgoing
cLSTM weights from input series $j$ to output component
series $i$, standardized by the maximum such edge weight
associated with the cLSTM for series $i$. Edges associated
with more than one degrees of freedom (angle directions
for the same body part) are averaged together. Finally, to
aid visualization, we further threshold edge weights of magnitude $0.01$ and
below.

\begin{figure}[ht]
\centering
\includegraphics[width=1.0\linewidth]{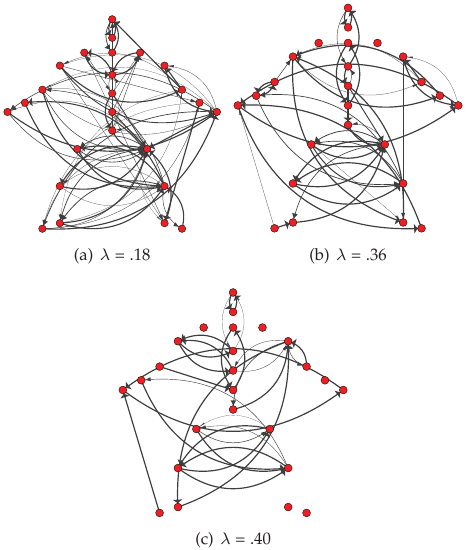}
\caption{Nonlinear Granger causality graphs inferred from the human
MoCap data set using the regularized cLSTM model. Results are
displayed for a range of $\lambda$ values. Each node corresponds to one location
on the body.}
\label{Fig9}
\end{figure}

The resulting estimated graphs are displayed in Figure \ref{Fig9}
for multiple values of the regularization parameter, $\lambda$. While
we present the results for multiple $\lambda$, one may use cross
validation to select $\lambda$ if one graph is required. To interpret
the presented skeleton plots, it is useful to understand the
full set of motion behaviors exhibited in this data set. These
behaviors are depicted in Figure \ref{Fig8}, and include instances
of \textit{jumping jacks, side twists, arm circles, knee raises, squats,
punching}, various forms of \textit{toe touches}, and \textit{running in place}.
Due to the extremely limited data for any individual behavior,
we chose to learn interactions from data aggregated over the
entire collection of behaviors. In Figure \ref{Fig9}, we see many
intuitive learned interactions. For example, even in the more
sparse graph (largest $\lambda$) we learn a directed edge from right
knee to left knee and a separate edge from left knee to
right. This makes sense as most human motion, including
the motions in this dataset involving lower body movement,
entail the right knee leading the left and then vice versa. We
also see directed interactions leading down each arm, and
between the hands and toes for toe touches.

\section{Conclusion}

We have presented a framework for nonlinear Granger causality selection using regularized neural network models of time series. To disentangle the effects of the past of an input series on the future of an output, we model each output series using a separate neural network. We then apply both the component multilayer perceptron (cMLP) and component long-short term memory (cLSTM) architectures, with associated sparsity promoting penalties on incoming weights to the network, and select for Granger causality. Overall, our results show that these methods outperform existing Granger causality approaches on the challenging DREAM3 data set and discover interpretable and insightful structure on a human MoCap data set.

Our work opens the door to multiple exciting avenues for future work. While we are the first to use a hierarchical lasso penalty in a neural network, it would be interesting to also explore other types of structured penalties, such as tree structured penalties \cite{kim2010tree}.

Furthermore, although we have presented two relatively simple approaches, based off MLPs and LSTMs, our general framework of penalized input weights easily accommodates more powerful architectures. Exploring the effects of multiple hidden layers, powerful recurrent and convolutional architectures, like clockwork RNNs \cite{koutnik2014clockwork}, and dilated causal convolutions \cite{oord2016wavenet}, open up a wide range of research directions and the potential to detect long-range and complex dependencies. Further theoretical work on the identifiability of Granger non-causality in these more complex network models becomes even more important.

\begin{figure}[ht]
\centering
\includegraphics[width=0.95\linewidth]{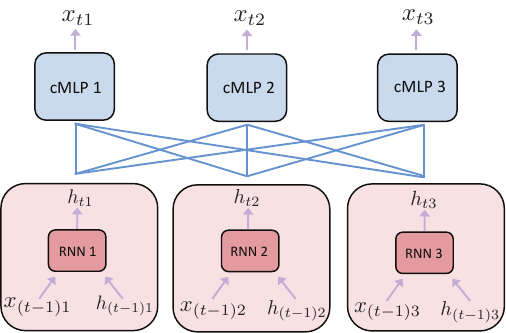}
\caption{Proposed architecture for detecting nonlinear Granger causality that combines aspects of both the cLSTM and cMLP models. A separate hidden representation, $h_{tj}$ , is learned for each series $j$ using an RNN. At each time point, the hidden states are each fed into a sparse cMLP to predict the individual output for each series $x_{ti}$. Joint learning of the whole network with a group penalty on the input weights of the individual cMLPs would allow the network to share information about hidden features in each $h_{tj}$ while also allowing interpretable structure learning between the hidden states of each series and each output.}
\label{Fig10}
\end{figure}

Finally, while we consider sparse input models, a different sparse output architecture would use a network, like an RNN, to learn hidden representations of each individual input series, and then model each output component as a sparse nonlinear combination across the hidden states of all time series, allowing a shared hidden representation across component tasks. A schematic of the proposed architecture that combines ideas from our cMLP and cLSTM models is
shown in Figure \ref{Fig10}.

\appendices
\section{Model Ablations}

We ran two ablation studies to understand factors that influence our methods' performance. First, we tested the cMLP and cLSTM with different numbers of hidden units on the Lorenz-96 data. Table~\ref{Table4} shows the AUROC results from a single run for two datasets with forcing constants $F \in (10, 40)$ and time series length $T = 1000$, using different numbers of hidden units, $H \in (5, 10, 25, 50, 100)$. The results reveal that both models are robust to a small number of hidden units, but that their performance improves with larger values of $H$. These findings suggest that overparameterization can help with the nonconvex optimization objective, leading to solutions that achieve high predictive accuracy while minimizing the penalty from the sparsity-inducing regularizer.

Next, we tested three approaches for optimizing our penalized objectives (Equations
8 
and
15). 
We compared standard gradient descent with Adam \cite{kingma2014adam} to proximal gradient descent (ISTA) \cite{parikh2014proximal} and proximal gradient descent with a line search (GIST) \cite{gong2013general} on the Lorenz-96 data with $T = 1000$ time points. Table~\ref{Table5} displays AUROC results across five initializations for two forcing constants $F \in (10, 40)$, using the cMLP with $H = 10$ hidden units. The results show that the three methods lead to similar results for both $F = 10$ and $F = 40$, although we did not compare the optimizers in other scenarios, e.g., with lower $T$ values or with the cLSTM.

Among these optimization approaches, Adam is fastest due to its adaptive learning rate, but it requires a parameter for thresholding the resulting weights (while the proximal methods lead to exact zeros). In contrast, GIST guarantees convergence to a local minimum and is less sensitive to the learning rate parameter, but it is also considerably slower than Adam and ISTA. We therefore use standard proximal gradient descent, or ISTA, in the remainder of our experiments, because it leads to exact zeros while being more efficient than GIST. In practice, this means running Algorithm~1 
or Algorithm~4 
using a fixed learning rate $\gamma$ rather than determining it by a line search.

\begin{table}[!t]
  \caption{AUROC comparisons for the cMLP and cLSTM as a function of the number of hidden units $H$ for simulated Lorenz-96 data. Results are calculated using a single run.}
  \label{Table4}
  \centering 
  \begin{tabular}{lcccc}\toprule
    Model & \multicolumn{2}{c}{cMLP} & \multicolumn{2}{c}{cLSTM} \\
    \cmidrule(r){1-1} \cmidrule(r){2-3} \cmidrule(r){4-5}
    $F$ & 10 & 40 & 10 & 40 \\
    \midrule
    $H$ = 5 & 96.5 & 91.0 & 91.9 & 86.9 \\
    $H$ = 10 & 98.0 & 94.0 & 94.5 & 91.5 \\
    $H$ = 25 & 98.4 & 94.3 & 95.6 & 92.3 \\
    $H$ = 50 & 98.3 & 94.4 & 95.7 & 93.8 \\
    $H$ = 100 & 98.5 & 94.5 & 95.7 & 95.2 \\
    \bottomrule
  \end{tabular}
\end{table}

\section{Baseline Methods}

The IMV-LSTM uses an attention mechanism to highlight the model's dependence on different parts of the input \cite{guo2019exploring}. We train a separate IMV-LSTM model to predict each time series using all the time series as inputs, using the ``IMV-Full'' variant \cite{guo2019exploring}, and we use the attention weights from the trained models to infer Granger causal relationships. Similar to the original work \cite{guo2018interpretable}, we record the empirical mean of the attention values for each input time series for each model, and we construct a $p \times p$ matrix of these values for the separate IMV-LSTMs. We then sweep over a range of threshold values to determine the most influential inputs for each IMV-LSTM, and we trace out an ROC curve from which we calculate AUROC values.

The LOO-LSTM baseline is based on the idea that withholding a highly predictive input should result in a decrease in predictive accuracy, a direction that has been explored for providing model-agnostic notions of feature importance \cite{lei2018distribution}, \cite{hooker2019please}. We begin by training separate LSTM models to predict each time series using all time series as inputs. We then train separate LSTM models to predict each time series $i$ using all inputs except time series $j$, and we record the increase in loss when the $j$th time series is withheld. Using the results, we construct a $p \times p$ matrix representing the differences in the loss, we sweep over a range of threshold values to determine the most influential inputs for each time series, and we trace out an ROC curve from which we calculate AUROC values.

\begin{table}[!t]
  \caption{AUROC comparisons between different optimization approaches for the cMLP with simulated Lorenz-96 data. Results are the mean across five initializations, with 95\% confidence intervals.}
  \label{Table5}
  \centering
  \begin{tabular}{lcc}\toprule
    $F$ & 10 & 40 \\
    \midrule
    GISTA & 98.0 $\pm$ 0.2 & 93.8 $\pm$ 0.3 \\
    ISTA & 98.0 $\pm$ 0.2 & 94.1 $\pm$ 1.9 \\
    Adam & 98.3 $\pm$ 0.1 & 95.1 $\pm$ 0.2 \\
    \bottomrule
  \end{tabular}
\end{table}

\ifCLASSOPTIONcompsoc
  \section*{Acknowledgments}
\else
  \section*{Acknowledgment}
\fi
AT, IC, NF and EF acknowledge the support of ONR Grant
N00014-15-1-2380, NSF CAREER Award IIS-1350133, and
AFOSR Grant FA9550-16-1-0038. AT and AS acknowledge
the support from NSF grants DMS-1161565 and DMS-1561814
and NIH grants R01GM114029 and R01GM133848.
\ifCLASSOPTIONcaptionsoff
  \newpage
\fi



\bibliographystyle{IEEEtran}
\bibliography{main}
\end{document}